\documentclass[lettersize,journal]{IEEEtran}
\usepackage{amsmath,amsfonts}
\usepackage{algorithmic}
\usepackage{algorithm}
\usepackage{array}
\usepackage[caption=false,font=normalsize,labelfont=sf,textfont=sf]{subfig}
\usepackage{textcomp}
\usepackage{stfloats}
\usepackage{url}
\usepackage{verbatim}
\usepackage{graphicx}
\usepackage{subcaption}
\usepackage{subfig}
\usepackage{wasysym}
\usepackage{ulem}
\usepackage{bm}

\usepackage{cite}

\def\R{\mathbb{R}}
\def\M{\mathbb{M}}
\def\E{\mathbb{E}}

\def\X{\mathcal{X}}

\def\g{\mathbf{g}}

\usepackage[numbers]{natbib}
\begin{document}

\title{Intrinsic-Hybrid Latent Diffusion Models for Generative Modeling on Unknown Manifolds}

\author{Yizhu Wang, Mu Niu, and Xiaochen Yang%
\thanks{Yizhu Wang, Mu Niu, and Xiaochen Yang are with the School of Mathematics and Statistics, University of Glasgow, Glasgow, U.K.}
}



\maketitle

\begin{abstract}
We introduce the Intrinsic Hybrid Latent Diffusion Model (ILDM), a generative framework that integrates probabilistic dimensionality reduction with geometry-aware diffusion on unknown manifolds. While diffusion models (DMs) have achieved state-of-the-art results in high-dimensional data synthesis, they rely on large training datasets and ignore intrinsic geometric structure. Latent diffusion models (LDMs) address the high dimensionality by learning a latent space, but they typically impose a Euclidean structure, failing to capture the underlying manifold geometry, especially problematic in data-sparse regimes. ILDM addresses these limitations by interpreting the latent space as a chart of an unknown Riemannian manifold, with geometry and uncertainty quantified through a probabilistic decoder. The forward process is a hybrid diffusion that switches between Riemannian and Euclidean dynamics based on local uncertainty, where the Riemannian component is governed by a probabilistic metric tensor derived from the decoder. To learn the generative dynamics, we introduce an approximate denoising score matching method tailored to the hybrid diffusion setting, enabling a backward process defined by hybrid Langevin dynamics. Experiments on COIL-100, MNIST, and cardiac MRI datasets demonstrate that ILDM significantly improves generation quality, achieving lower FID and LPIPS scores compared to standard diffusion and latent diffusion models.
\end{abstract}


\section{Introduction}
\IEEEPARstart{G}{enerative} modeling via diffusion processes has emerged as a transformative framework for synthesizing high-dimensional data, achieving state-of-the-art performance in tasks such as image generation \citep{ho2020denoising,song2019generative}. Recent progress has significantly expanded the applicability of diffusion-based models, enabling breakthroughs in diverse domains such as high-fidelity image synthesis \citep{karras2023elucidating} and text-to-video generation \citep{singer2023text2video}. Central to this paradigm are score-based generative models \citep{vincent2011connection,sohl2015deep,song2020score}, which learn to reverse predefined stochastic differential equations (SDEs) that gradually perturbs data into noise. Although recent advances \citep{song2020score} have unified the design of diffusion SDEs and improved sampling efficiency, these models usually require a large number of training data to generate good quality images, limiting their potential in sparse-data regimes. The manifold hypothesis \citep{fefferman2016testing} provides an important perspective on this issue: real-world data distributions often concentrate near low-dimensional manifolds embedded in high-dimensional ambient spaces. This hypothesis empirically holds for many datasets and has become the foundation of manifold learning. Leveraging this intrinsic structure could, in principle, simplify generative modeling by reducing ambient dimensionality. When the manifold geometry is known, various generative methods have been developed. Extrinsic approaches \citep{lin2019,huang2022riemannian} embed the manifold into a higher-dimensional Euclidean space and define diffusion in the ambient space. While computationally convenient, these methods depend on access to accurate embeddings. Intrinsic approaches avoid embeddings by directly modeling diffusion on the manifold. \citet{de2022riemannian} used Geodesic Random Walks and heat kernel estimation, while \citet{jo2024generative} proposed bridge-based processes using the logarithm map. However, these methods require explicit knowledge of manifold operators, such as the Laplace-Beltrami eigenbasis, limiting their use to manifolds with explicitly known structure.

In practice, however, the manifold geometry is often unknown. Latent diffusion models (LDMs) \citep{rombach2022high} address this challenge by learning a low-dimensional latent representation via a dimensionality reduction approach, such as the variational autoencoder (VAE)~\citep{kingma2019} or Gaussian process (GP) decoder~\citep{lawrence2005,lawrence2007,moreno2022}, and performing diffusion in this latent space. However, these approaches typically impose a Euclidean structure on the latent space, disregarding the true geometry of the underlying manifold.
Additionally, they use dimensionality reduction only to create a low-dimensional space, ignoring it in the subsequent forward and backward diffusion processes. As a result, two challenges arise:
i) \textit{geometric mismatch}, where the forward and backward diffusion processes assume Euclidean dynamics that fail to reflect the geometry and metric variations of the manifold; and ii) \textit{sparse data degradation}, where limited training data lead to increased reconstruction errors in regions of high mapping uncertainty, due to the lack of robustness of manifold learning methods beyond densely sampled area.

In this work, we propose intrinsic hybrid latent diffusion models (ILDM), a generative framework that integrates probabilistic dimensionality reduction with geometry-aware hybrid diffusion processes. ILDM is particularly effective in data-sparse settings, where conventional diffusion models often struggle. At the core of ILDM is a hybrid diffusion model designed to operate on latent spaces learned via pretrained decoders. We interpret the latent space as a chart of an unknown manifold equipped with Riemannian metric. This metric encodes both the intrinsic geometry and the uncertainty of the latent-to- data mapping, forming the basis for geometry-aware diffusion. The forward process evolves as a mixture of Riemannian and Euclidean Brownian motion, switching dynamically based on mapping uncertainty. To learn the generative dynamics, we estimate the time-dependent score function from simulated hybrid trajectories using an approximate denoising score matching objective. The backward process is formulated as a hybrid of Riemannian and Euclidean Langevin dynamics, guided by the learned scores and adjusted according to the Riemannian metric.

We evaluated ILDM on three benchmark datasets under sparse data conditions, the `Lucky Cat' object of COIL-100~\citep{nene1996}, left ventricle cardiac MRI scans~\citep{bernard2018deep,campello2020multi} and subsets of MNIST~\citep{lecun1998mnist}. ILDM consistently outperforms conventional score-based diffusion models(DM) and LDMs, achieving lower Fréchet Inception Distance (FID) and LPIPS scores.

\section{Concepts of Riemannian Geometry and Data Noise Manifold}
\label{sec:rieman}
\begin{figure}[t]
    \centering
\includegraphics[width=0.4\textwidth,height=0.2\textwidth]{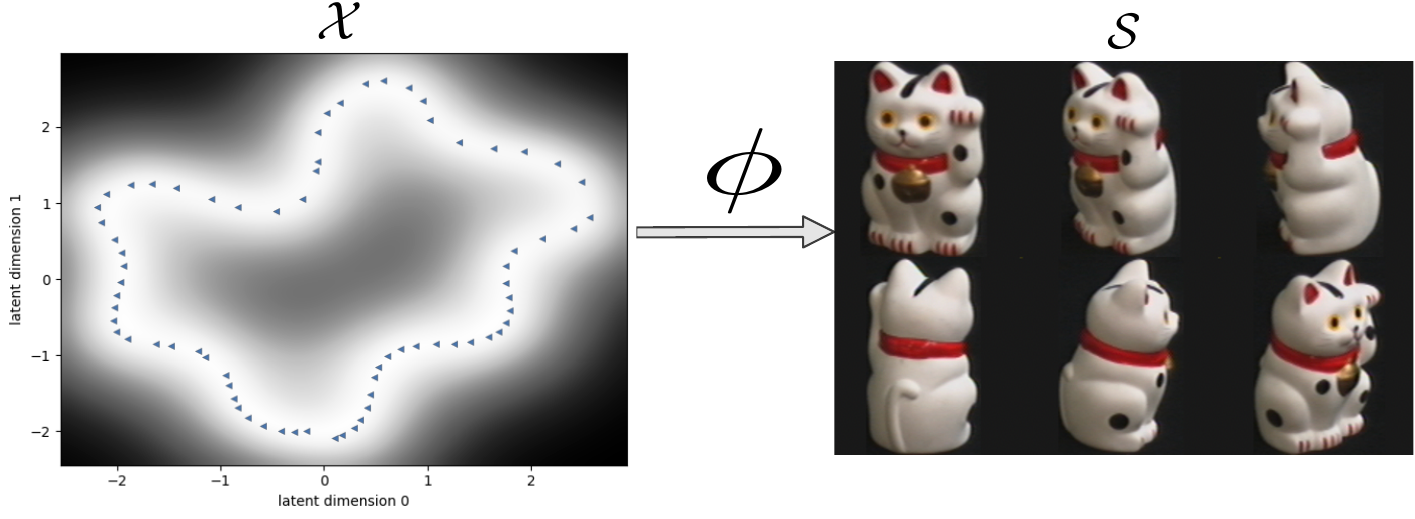}
    \caption{ 
    \label{fig:Coillatent} 
    \footnotesize{ Illustration of manifold parameterization for the COIL image dataset. The mapping $\phi:\mathcal{X}\rightarrow \mathcal{S}$ projects latent points (blue triangles in the left sub-figure) to their corresponding images (six examples shown in the right sub-figure). Background shading in the latent space  represents the uncertainty of $\phi$, quantified as the variance of the mapping, which increases with distance from observed data points.
    }}
\end{figure}

A widely used strategy for modeling high-dimensional data is to map it into a lower-dimensional latent space via a set of (potentially nonlinear) functions. Previous works~\citep{tosi2014metric, arvanitidis2019} have explored the geometric properties of probabilistic generative models for dimensionality reduction -- such as Gaussian process decoders (GP decoders) and VAEs. These models define a latent manifold embedded in high-dimensional space via a probabilistic mapping from the latent coordinates. As illustrated in Figure~\ref{fig:Coillatent} for the 'Lucky Cat' object of COIL dataset, latent points (blue triangles) are mapped to the image space through a mapping function $\phi$, revealing a compact and structured representation of the data. In this work, we adopt both GP decoders and VAEs for dimensionality reduction. These decoders are pretrained prior to training the diffusion models, and their parameters remain fixed throughout~\citep{rombach2022high}. Serving as a bridge between the image space and latent space, they enable diffusion processes to operate efficiently within the lower-dimensional latent domain. Additional descriptions of GP decoders and VAEs are provided in the Appendix H. 

Let $s_i \in \R^p$ denote the observed data point (or image) in a high-dimensional space (or image space). We define $\mathcal S = \{ s_i |  i=1,\cdots, n \}$, where $\mathcal{S} \in \R^{ n \times p }$. To model the data, we introduce latent variables $x_i$, where $x_i \in \mathbb{R}^q$ with $q < p$. We have $\mathcal{X}= \{x_i | i= 1,\cdots, n \}$, where $\mathcal{X} \in \R^{n\times q}$. The latent variables $x_i$ are linked to the observed data $s_i$ via a probabilistic mapping:
\begin{align} \label{eqn:mapmodel}
s_{i} = \phi(x_i) + e_{i},
\end{align}
where $e_i$ is an independent Gaussian noise term, and $\phi$ is modeled by the decoder.

It is important to recognize that not all points in the latent space yield plausible reconstructions in the data space. Probabilistic decoders often extrapolate poorly in regions far from training data. In these regions, reconstructions are dominated by noise. Unlike the data manifold assumed in classical manifold learning approaches \citep{belkin2003laplacian,roweis2000nonlinear}, where every point corresponds to a valid high dimensional observation, in our setting, the latent space describes a \textit{data noise manifold}: a $q$-dimensional sub-manifold embedded in the $p$-dimensional data space, capturing both meaningful data structures and the surrounding regions dominated by noise. 

The latent space $\mathcal{X}$ serves as a chart for this manifold. Intuitively, a chart provides a locally distorted view of the underlying manifold. While distances and measurements are defined intrinsically on the manifold, they can be computed locally in the chart (or latent space) and then integrated to obtain global quantities. This leads naturally to the notion of a Riemannian metric $\g$: a smoothly varying, symmetric, positive-definite matrix that defines an inner product on the tangent space at each point of the manifold \citep{Lee13}. Let $\M$ be a $q$-dimensional  Riemannian manifold with the Riemannian metric $\g$. The metric quantifies how infinitesimal perturbations in the latent space are stretched or compressed when mapped to the data space via $\phi$, and thus governs the geometry-aware behavior of diffusion processes on the manifold.

Formally, let $\phi: \mathbb{R}^q \rightarrow \mathbb{R}^p$ be the mapping from latent to data space, and let $\mathcal{J}$ denote its Jacobian matrix. The Riemannian metric $\mathbf{g}$ is given by:
\begin{align} \label{eqn:deter_metric}
\mathbf{g} = \mathcal{J}^\top \mathcal{J}, \quad \text{with} \quad \mathcal{J}_{i,j} = \frac{\partial \phi^i}{\partial x^j},
\end{align}
where the superscript $i = 1, \dots, p$ indexes the data-space dimensions, and $j = 1, \dots, q$ indexes the latent dimensions. The metric $\g \in \mathbb{R}^{q \times q}$ encodes local geometric distortions introduced by the mapping $\phi$. For notational simplicity, we write
$\g$ without explicitly indexing its dependence on
$x$, though it is understood to be position-dependent in latent space.

When using GP decoders, the mapping $\phi$ is modeled as a multivariate Gaussian distribution, and so does the Jacobian \citep{Rasmussen2006}, $p(\mathcal{J} | \X , S ) = \prod_{l=1}^p \mathcal{N} ( \mu_{J}^l, \Sigma_{J}  )$. The corresponding metric follows a noncentral Wishart distribution \citep{anderson1946}. We use the mean of the noncentral Wishart distribution to define $\g$ as
\begin{align} \label{eqn:GP_metric}
\g = \E( \mathcal{J}^\top) \E( \mathcal{J ) } + p\Sigma_J.
\end{align}
Further details and expressions for $\mu_{J}^l$ and $\Sigma_{J}$ are provided in Appendix H-A. 
For VAEs, where the uncertainty of the mapping is not explicit, we can use the formulation in Equation~(\ref{eqn:deter_metric}) to construct the metric. Additional details on metric construction for VAEs are given in Appendix H-B.

\section{Intrinsic Hybrid Latent Diffusion Model }
Following the construction of the mapping $\phi$ and the associated Riemannian metric 
$\g$ on the latent space, we now turn to the task of defining diffusion processes that are consistent with the learned manifold structure.  In particular, we aim to model stochastic dynamics that respect the intrinsic geometry of the manifold and adapt to varying levels of uncertainty across different regions of the latent space.

In standard score-based generative modeling, data are progressively perturbed across multiple noise scales, and the evolution of the perturbed distributions is governed by SDEs in Euclidean space. Here, we generalize this idea by considering diffusion processes that evolve on the learned manifold. However, as discussed earlier, the estimated metric $\g$ becomes unreliable in regions far from the training data, where the uncertainty of the mapping $\phi$ is high. To address this, we introduce a hybrid diffusion approach: in regions of low uncertainty, we simulate manifold-based stochastic processes; in high-uncertainty regions, we switch to standard Euclidean dynamics, where manifold structure is less informative. This allows the generative process to naturally account for geometry and uncertainty of the unknown manifold.

\begin{figure*}[t]
    \centering
\includegraphics[width=0.9\textwidth,height=0.3\textwidth]{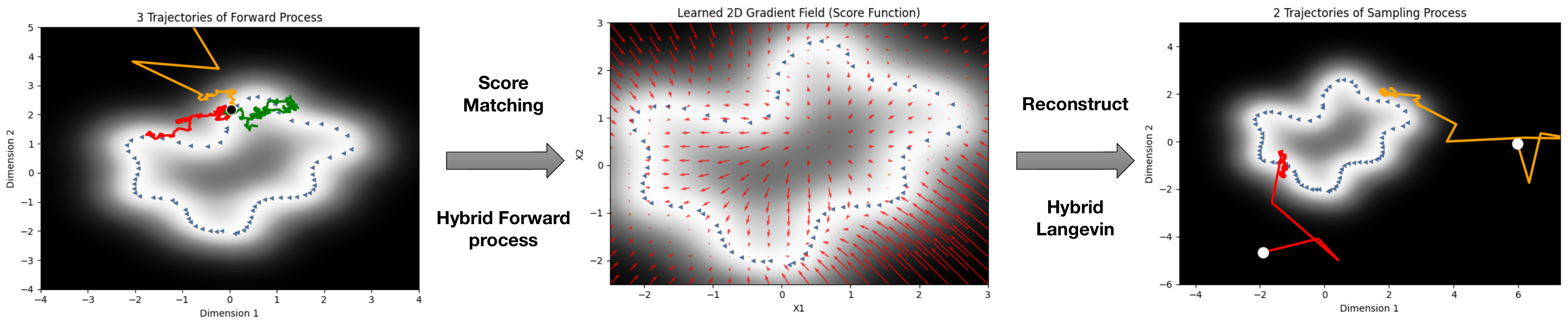}
    \caption{ \label{fig:overview} 
    \footnotesize{ 
    {Score-based generative model on the data noise manifold.
    The blue triangles in each sub-figure represent the data points in the latent space. 
    The left sub-figure shows three Brownian motion~(BM) paths in the latent space. Starting from the black circle, the red and green paths follow Riemannian BM, while the yellow path initially follows Riemannian BM but switches to Euclidean BM when it enters the high uncertainty (dark) region. Once the BM paths are simulated, the vector field of score function ($\nabla_x \log p(x(t))$) is learned as described in Section \ref{sec:score-estimation}. The gradients at some grid points are shown in middle sub-figure, and they clearly point toward the data points (blue triangles). The learned scores are then used to construct the backward process. The right panel shows two backward paths, which start from white circles and converge toward the data distribution. 
    }}
    }
\end{figure*}

In the following, we formalize the forward and backward diffusion processes under this hybrid approach. We also describe how score estimation is adapted to this setting, where closed-form transition densities are not available. An overview of the proposed framework is provided in Figure \ref{fig:overview}.

\subsection{Forward process}
\label{sec:forward}
We define the forward process $\{x(t)\}_{t=0}^T $ on the unknown manifold, indexed by a continuous time variable $t\in[0,T]$. We denote the probability density of $x(t)$ by $p_t(x)$. The initial state follows $x(0) \sim p_0$, where $p_0$ represents the data distribution, and we have access to an i.i.d. dataset of samples. The terminal state is distributed as $x(T) \sim p_T$, where $p_T$ serves as the prior distribution, from which we can generate samples. The transition density from $x(t_i)$ to $x(t_j)$is denoted as $p_{t_i t_j}(x(t_i)| x(t_j))$. The forward process alternates between two types of diffusion mechanisms, depending on the uncertainty in $\phi$. In confident regions, we simulate Riemannian Brownian motion (BM) on manifold. Conversely, in high-uncertainty regions, we transition to the BM in Euclidean space, a special case of variance-exploding SDEs as introduced in \cite{song2020score}. 

Given the estimate of the metric $\g$ as in Section \ref{sec:rieman}, the BM on a Riemannian manifold in a local coordinate system is formulated as a system of SDEs in the It\^{o} form \citep{hsu1988,hsu2008}:  
\begin{align}
\label{eqnBMhsu}
dx^i(t) = \frac{1}{2}G^{-1/2} \sum^{q}_{j=1}\frac{\partial}{\partial x^j} \left(  {\g^{-1}}_{ij}G^{1/2} \right) dt + \left( \g^{-1/2} db(t)\right)_i
\end{align}
where $G$ is the determinant of $\g$, $b(t)$ represents an independent BM in the Euclidean space, and $i$ is the $i$-th dimension of the latent space. The discretized version of the SDEs is derived from the Euler-Maruyama method \citep{kloeden1992,lamberton2007}:
\begin{align}
\label{eqn:disBM}
x^i(t) &= x^i(t-\Delta t) 
    + \frac{1}{2} \sum_{j=1}^{q} 
        \Biggl( - \g^{-1} \frac{\partial \g}{\partial x^j} \g^{-1} \Biggr)_{ij} 
        \Delta t \notag \\
& \hspace{4em}
    + \frac{1}{4} \sum_{j=1}^q (\g^{-1})_{ij} 
        \, \mathrm{tr} \Biggl( \g^{-1} \frac{\partial \g}{\partial x^j} \Biggr) 
        \Delta t 
    + \bigl( \g^{-1/2} db(t) \bigr)_i \notag \\
&= \mu \bigl( x^i(t-\Delta t), \Delta t \bigr) 
    + \bigl( \sqrt{\Delta t} \, \g^{-1/2} z_t \bigr)_i
\end{align}

where $\Delta t$ is the diffusion time of each step of the BM simulation, $tr(\cdot)$ is the trace operator, $z_t$ represents a $q$-dimensional Gaussian-distributed random vector, and $\mu(x(t{-}\Delta t), \Delta t)$ denotes the drift term consisting of the first three deterministic components in the discretized update rule. Notably, both the drift and diffusion terms depend on $\g$, which varies with position $x$ in the latent space. The estimated metric $\g$ is only reliable in regions of low mapping uncertainty.

{\it Hybrid Switching.} The reliability of the Riemannian metric is determined by the predictive variance \( \sigma^2(x) \) of the mapping $\phi(x)$. As a point in the latent space moves away from the data, the variance increases and eventually plateaus at the maximum value $\sigma^2_{max}$. In the case of GP decoder, \( \sigma_{\max}^2 \) corresponds to the model's prior variance. To determine whether the forward process should follow manifold-based or Euclidean dynamics, we introduce a switching threshold based on this maximum predictive variance. Specifically, we introduce a scaling parameter \( \alpha \in (0, 1) \) and set the switching threshold $\sigma^2_\text{thresh}$ as the fraction of the maximum variance: $\sigma_{\text{thresh}}^2 = \alpha \, \sigma_{\max}^2$. When $\sigma^2(x) < \sigma^2_{\text{thresh}}$, the forward process follows Riemannian BM on the learned manifold; otherwise, it switches to Euclidean BM. This allows the model to leverage the geometric structure where $\phi$ is confident, while permitting more flexible exploration in high-uncertainty regions. There is no closed-form solution for this hybrid forward process. $p_{0t}(x_t|x_0)$ is not tractable.
Nevertheless, the hybrid SDE can be simulated numerically using schemes such as Euler-Maruyama. Figure~\ref{fig:overview} (left) illustrates three representative hybrid forward trajectories initialized from the same starting point (black circle). While two trajectories (red, green) remain as Riemanian BM throughout, the third (yellow) switches to Euclidean dynamics upon entering a high-uncertainty region. The complete simulation procedure is detailed in Algorithm~\ref{alg:forward}.

An ablation study examining the impact of $\alpha$ on generative quality is provided in Appendix B. 
We empirically tested $\alpha \in \{0.3,0.4,0.5 \}$, corresponding to thresholds at $30\%-50\%$ of $\sigma^2_{\max}$, respectively. Across all settings, the ILDM-generated images consistently achieved lower FID scores compared to those produced by the standard LDM, indicating improved sample quality. In practice, the value of 
$\alpha$ can be treated as a tunable hyperparameter and selected via cross-validation. While this work specifically considers BM in $\R^q$ for high uncertainty regions,  our framework is flexible and could incorporate alternative SDEs in high uncertainty region, such as those described in \cite{song2020score}. For VAEs, a related construction of Riemannian metrics and uncertainty-driven dynamics is provided in Appendix H-B. 

\begin{algorithm}[H]
\caption{Forward Diffusion Process on Unknown Manifold}
\label{alg:forward}
\begin{algorithmic}
\STATE 
\STATE \textbf{Input:} 
       Learn metric $\g$ from point cloud 
       $\mathcal{S} = \{ s_i \mid i = 1,\ldots,n_d \}$;  
       $n_f$: Number of starting points;  
       $n_{BM}$: Number of trajectories per starting point;  
       $n_{t_f}$: Number of diffusion steps;  
       $\Delta t_f$: Step size;  
       $\sigma^2_{\text{thresh}}$: Switching variance threshold.
\STATE \textbf{Output:} Simulation trajectories $x$.
\STATE 
\STATE \textbf{for } $i = 1, \ldots, n_f$ \textbf{ do}
\STATE \hspace{0.5cm} \textbf{for } $j = 1, \ldots, n_{BM}$ \textbf{ do}
\STATE \hspace{1.0cm} \textbf{for } $l = 1, \ldots, n_{t_f}$ \textbf{ do}
\STATE \hspace{1.5cm} \textbf{if } $\mathrm{Var}\!\left( \phi(x_{i,j}(l{-}1)) \right) 
                                   < \sigma^2_{\text{thresh}}$ \textbf{ then}
\STATE \hspace{2.0cm} $q\!\left( x_{i,j}(l) \mid x_{i,j}(l{-}1) \right) \gets$ 
\STATE \hspace{2.5cm} $\mathcal{N}\!\Big(
                x_{i,j}(l) \, \Big| \,
                \mu\!\left( x_{i,j}(l{-}1), \Delta t_f \right),$
\STATE \hspace{2.5cm} $ \Delta t_f \, \g^{-1} \Big)$ {\scriptsize (Eq.~\eqref{eqn:disBM})}
\STATE \hspace{1.5cm} \textbf{else}
\STATE \hspace{2.0cm} $q\!\left( x_{i,j}(l) \mid x_{i,j}(l{-}1) \right) \gets$
\STATE \hspace{2.5cm} $\mathcal{N}\!\Big(
              x_{i,j}(l) \, \Big| \,
              x_{i,j}(l{-}1), \Delta t_f \Big)$
\STATE \hspace{1.5cm} \textbf{end if}
\STATE \hspace{1.0cm} \textbf{end for}
\STATE \hspace{0.5cm} \textbf{end for}
\STATE \textbf{end for}
\STATE \textbf{return} $x$
\end{algorithmic}
\end{algorithm}

\subsection{Estimating Scores via Time-Dependent Score Matching}
\label{sec:score-estimation}

To enable generative sampling in our ILDM, we require estimating the time-dependent score function \( \nabla_x \log p_t(x) \), where \( p_t(x) \) is the density of the forward process at time \( t \). This score function is approximated by a neural network \( s_{\theta}(x, t): \mathbb{R}^q \times [0,T] \to \mathbb{R}^q \), which we train using variants of score matching \citep{hyvarinen2005estimation,ronneberger2015u}. In standard LDMs, where the forward process operates in Euclidean latent space and admits a closed-form transition density, denoising score matching (DSM) can be directly applied to supervise the score network using known perturbation distributions \citep{vincent2011connection,song2019generative}. However, in ILDM, the forward process is governed by a hybrid SDEs on an unknown manifold. The transition density \( p_{0t}(x_t|x_0) \) is intractable, precluding the direct use of standard DSM.

To address this challenge, we consider two approaches based on simulated forward trajectories: a sliced score matching (SSM) objective adapted from \citet{song2019generative}, and a novel \emph{approximate denoising score matching (ADSM)} objective developed in this work. Both methods rely on simulating the forward SDEs paths (Algorithm~\ref{alg:forward}) to obtain samples from the transition distribution, which are then used to train the score network.

In ADSM, for each $x_0 \sim p_0$, we simulate $n_{BM}$ paths of the hybrid forward SDEs, each with $n_{tf}$ steps (see Algorithm \ref{alg:forward} ). We approximate the intractable transition density \( p_{0t}(x_t \mid x_0) \) by a Gaussian \( \mathcal{N}(\tilde{\mu}_t, \tilde{\tau}_t^2) \), where \( \tilde{\mu}_t \) and \( \tilde{\tau}_t^2 \) denote the sample mean and variance of simulated trajectories at time $t$, stating from that specific $x_0$. This yields an approximation of the score function:
$
\nabla_x \log p_{0t}(x_t \mid x_0) \approx \frac{x_t - \tilde{\mu}_t}{\tilde{\tau}_t^2},
$
which we use to define the following training objective: 
\begin{align}
\label{eqn:approxSM}
\theta &= \arg\min_{\theta} \mathbb{E}_t \Biggl\{ 
    \lambda(t) \mathbb{E}_{x(0)} \mathbb{E}_{x(t) \mid x(0)} 
    \Biggl[ 
        \biggl\| s_\theta(x_t, t) \cdot \tilde{\tau}_t  \notag \\
& \hspace{4em}
        - \frac{x_t - \tilde{\mu}_t}{\tilde{\tau}_t} 
        \biggr\|^2 
    \Biggr] 
\Biggr\},
\end{align}
where \( \lambda(t): [0,T] \to \mathbb{R}^+ \) is a time-dependent weighting function. Here, $\lambda(t)$ is computed using the marginal variance of $x_t$, estimated over all simulated trajectories at time $t$, aggregated across all $x_0$. This global weighting scheme stabilizes training and ensures consistent scaling across time steps, in line with the noise-aware weighting strategy in \citet{song2019generative}. The score network $s_{\theta} (x,t)$ is implemented as a U-Net with convolution layers (appendix G). 

Alternatively, SSM also offers a density-free approach to score estimation  based on random directional projections. Following \citet{song2020score}, we define the objective as:
\begin{align}
\label{eqn:sliceSM}
    \theta &= \arg\min_{\theta} \mathbb{E}_t \Biggl\{ 
        \lambda(t) \mathbb{E}_{x(0)} \mathbb{E}_{x(t)} \mathbb{E}_{v\sim p_v} 
        \Biggl[ \frac{1}{2} \bigl\| s_{\theta}(x(t),t) \bigr\|_2^2  \notag \\
    & \hspace{4em}
        + v^\top \nabla_x s_{\theta}(x(t),t) v 
        \Biggr] \Biggr\},
\end{align}
where \( v \sim p_v \) is a random direction vector drawn from a standard multivariate Gaussian distribution. While SSM circumvents the need for transition densities, it can suffer from slow convergence in high dimensions due to its reliance on random projections, which yield high-variance gradient estimates. 

In contrast, ADSM yields more stable and efficient optimization. Despite assuming a Gaussian approximation to the conditional distribution, our experiments demonstrate that the estimated score field closely aligns with that from SSM. In both methods, evaluating the score network $s_{\theta}(x_t,t)$ at grid points in latent space yields a vector field, which can be interpreted the gradient of the log-density. Throughout, we use the terms score function and vector field interchangeably. Additional comparisons of vector field visualizations and training stability for Approximate DSM and SSM are provided in Appendix B. 

\begin{figure*}[b]
    \centering
    \begin{minipage}[t]{0.48\textwidth}
        \centering
        \includegraphics[width=0.95\textwidth,keepaspectratio]{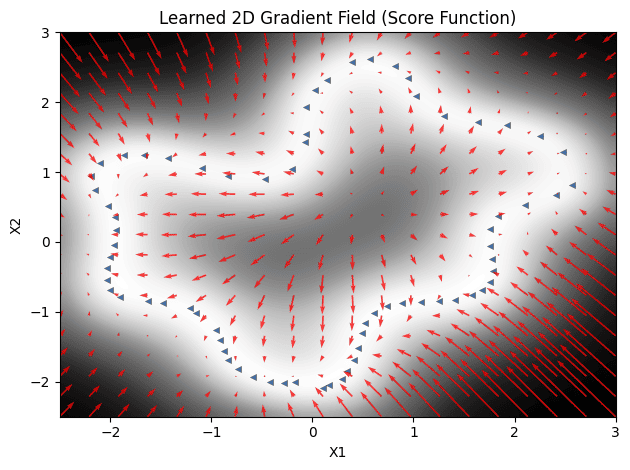}
        \vspace{0.3em}
        \makebox[\linewidth][c]{\footnotesize (a) Vector field of ILDM}
    \end{minipage}
    \hfill
    \begin{minipage}[t]{0.48\textwidth}
        \centering
        \includegraphics[width=0.95\textwidth,keepaspectratio]{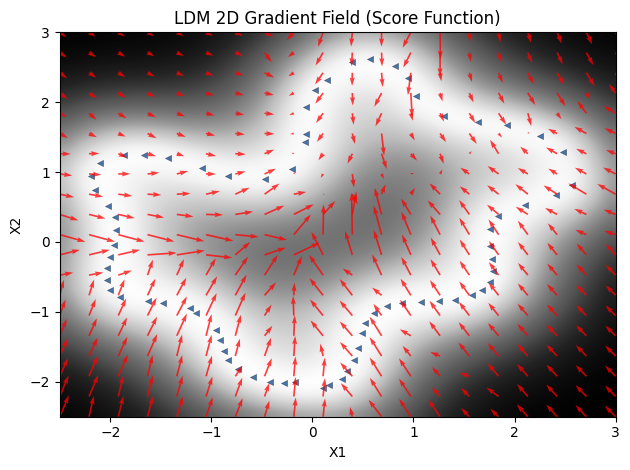}
        \vspace{0.3em}
        \makebox[\linewidth][c]{\footnotesize (b) Vector field of LDM}
    \end{minipage}

    \vspace{0.8em}

    \caption{
    \makebox[\textwidth][c]{\footnotesize Comparison of vector field estimates from ILDM (a) and LDM (b).}
    }
    \label{fig:VF}
\end{figure*}

To illustrate the impact of geometry-aware score estimation, we compare ILDM (via ADSM) with the standard LDM using the COIL image dataset. Figure~\ref{fig:VF} shows the learned vector field in latent space. In ILDM (left panel), the estimated vector field consistently points toward data distribution (blue triangles), accurately capturing the local geometry of the data distribution. In contrast, the LDM vector field (right panel) predominantly points toward the center of the latent space, where no data exist, revealing its failure to capture the underlying manifold structure.

\subsection{Backward process}
After training the time-dependent score model
$s_\theta(x_t,t)$, we construct the backward process to draw samples from the data distribution $p_0$. The backward process is designed as a hybrid of Euclidean and Riemannian Langevin dynamics \citep{girolami2011riemann,chen2014stochastic}, adapting to the uncertainty in $\phi$.

Starting from $x(t)\sim p_T$, we simulate the backward process to obtain samples $x(0)\sim p_0$. In regions of high mapping uncertainty, the process follows Euclidean Langevin dynamics. Since we are running the process backward in time (time flows from $T$ to 0), the corresponding SDE is given by
\begin{equation}
    dx(t) = -\frac{1}{2} \nabla_x \log(p(x(t)) ) dt + d{b}(t),
\end{equation}
where $b(t)$ is a independent Euclidean BM and $dt$ is an infinitesimal negative timestep. Replacing the score $\nabla_x \log p(x(t))$  with the learned estimator $s_\theta(x_t,t)$, we can simulate the SDE with numerical solver, which discretizes the SDE using finite time steps and small Gaussian noise.

In low uncertainty regions, we employ Riemannian Langevin dynamics \citep{girolami2011riemann}. The corresponding backward-time SDE is given by
 \begin{equation} 
 \label{eqn:mLan_main}
 dx(t) = -\frac{1}{2} \g^{-1} \nabla_x \log p_t(x(t)) dt + d\tilde{ \mathcal{B}}(t), \end{equation}
where $\g^{-1} \nabla_{x} \log p_t(x(t))$ is the natural gradient \citep{amari2000methods}, and $\tilde{ \mathcal{B}}(t)$ is the Brownian motion on manifolds. If we expand Equation \eqref{eqn:mLan_main} for the $i$-th coordinate of the latent space and replace $\nabla_x \log p_t(x(t))$ with the estimated score $s_{\theta}(x_t,t)$, we obtain the following SDEs:
\begin{align}
\label{eqn:rlan}
  dx^i(t) =&  -\left( \frac{1}{2} \g^{-1} s_{\theta}(x_t,t) dt \right)_i + \frac{1}{2} G^{-1/2} \sum_{j=1}^{q} \frac{\partial}{\partial x^j} \left( {\g^{-1}}_{ij} G^{1/2} \right) dt \nonumber \\ 
  & +\left(\g^{-1/2} db(t)\right)_i.
\end{align}
As with the forward process, the hybrid backward SDE is simulated using Euler-Maruyama, with a switching rule based on the variance of mapping function. The variance threshold $\sigma^2_\text{thresh}$ is defined as in Section \ref{sec:forward}. The discretization details of equation \eqref{eqn:rlan} are given in Appendix A. 

Figure~\ref{fig:overview} (right panel) illustrates example trajectories of the backward hybrid process. Two paths are initialized from points in high-uncertainty regions (dark), far from the data distribution. The trajectories initially follow the Euclidean Langevin diffusion and move with large steps. As they enter the low uncertainty region (white), the dynamics switch to Riemmanian Langevin updates and converge towards the data distribution. The complete backward sampling procedure is presented in Algorithm \ref{alg:backward}.

\begin{algorithm}[H]
\caption{Backward Langevin Process on Unknown Manifold}
\label{alg:backward}
\begin{algorithmic}
\STATE 
\STATE \textbf{Input:} 
       $n_b$: Number of backward points;  
       $n_{La}$: Number of trajectories;  
       $n_{t_b}$: Number of backward steps;  
       $\Delta t_b$: Step size;  
       $\sigma^2_{\text{thresh}}$: Variance threshold;  
       $\mathcal{G}$: Riemannian metric;  
       $x_T$: Terminal states.
\STATE \textbf{Output:} Reconstructed $\tilde{x}$.
\STATE 
\STATE Initialize backward starting points 
       $x_{i,j}(n_{t_b}) \leftarrow x_T$.
\STATE \textbf{for } $i = 1, \ldots, n_b$ \textbf{ do}
\STATE \hspace{0.5cm} \textbf{for } $j = 1, \ldots, n_{La}$ \textbf{ do}
\STATE \hspace{1.0cm} \textbf{for } $l = n_{t_b}, \ldots, 1$ \textbf{ do}
\STATE \hspace{1.5cm} \textbf{if } $\mathrm{Var}\!\left( \phi(x_{i,j}(l)) \right) 
                                   < \sigma^2_{\text{thresh}}$ \textbf{ then}
\STATE \hspace{2.0cm} $q\!\left( x_{i,j}(l{-}1) \mid x_{i,j}(l) \right) \gets$ 
\STATE \hspace{2.5cm} $\mathcal{N}\!\Big( x_{i,j}(l{-}1) \, \Big| \,
              \tilde{\mu}\!\left( x_{i,j}(l), \Delta t_b \right),$
\STATE \hspace{2.5cm} $ \Delta t_b \, \mathcal{G}^{-1} \Big)$ {\scriptsize (Eq.~12)}
\STATE \hspace{1.5cm} \textbf{else}
\STATE \hspace{2.0cm} $q\!\left( x_{i,j}(l{-}1) \mid x_{i,j}(l) \right) \gets$
\STATE \hspace{2.5cm} $\mathcal{N}\!\Big( x_{i,j}(l{-}1) \, \Big| \,
              \mu\!\left( x_{i,j}(l), \Delta t_b \right),$
\STATE \hspace{2.5cm} $\Delta t_b \Big)$ {\scriptsize (Eq.~13)}
\STATE \hspace{1.5cm} \textbf{end if}
\STATE \hspace{1.0cm} \textbf{end for}
\STATE \hspace{0.5cm} \textbf{end for}
\STATE \textbf{end for}
\STATE \textbf{return} $\tilde{x}$
\end{algorithmic}
\end{algorithm}

\subsection{Controllable generation with classifier-free guidance}
We can extend our framework to class conditional generation using classifier-free guidance. Following the approach of \citet{ho2022classifier}, we train a conditional score network  \( s_\theta(x,t,c) \) alongside an unconditional score network \( s_\theta(x,t) := s_\theta(x,t,c=\emptyset) \), both implemented using a shared U-Net architecture \citep{ronneberger2015u}. The networks are jointly optimized using the approximate DSM objective (Equation~\eqref{eqn:approxSM}). 
During training, the conditioning input \( c \) is randomly set to null with probability \( p_\text{uncond} \), allowing the model to simultaneously learn both conditional and unconditional score functions. As demonstrated in \citet{ho2022classifier, dhariwal2021diffusion}, small values of \( p_\text{uncond} \in \{0.1,0.2\} \) are sufficient for effective guidance without compromising conditional model performance. We adopt a similar setting and find it adequate for high-quality guided sampling within our ILDM framework. Detailed algorithms for training the joint score network are provided in Appendix C. 

\begin{figure*}[h]
    \centering
    \begin{minipage}[t]{0.45\textwidth}
        \centering
        \includegraphics[width=0.9\textwidth,keepaspectratio]{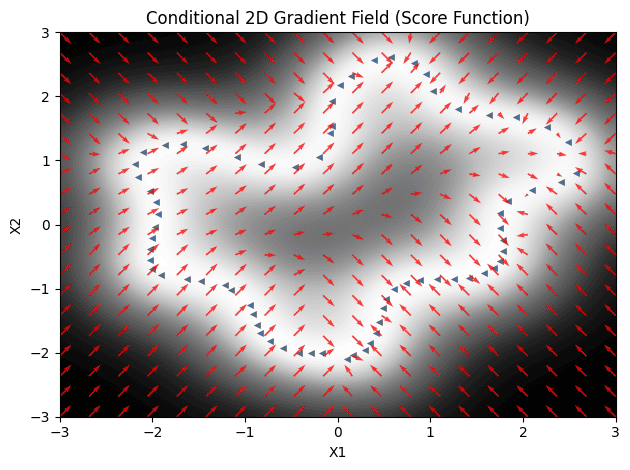}
        \vspace{0.3em}
        \makebox[\linewidth][c]{\footnotesize (a) Condition vector field estimated by ILDM}
    \end{minipage}
    \hfill
    \begin{minipage}[t]{0.45\textwidth}
        \centering
        \includegraphics[width=0.9\textwidth,keepaspectratio]{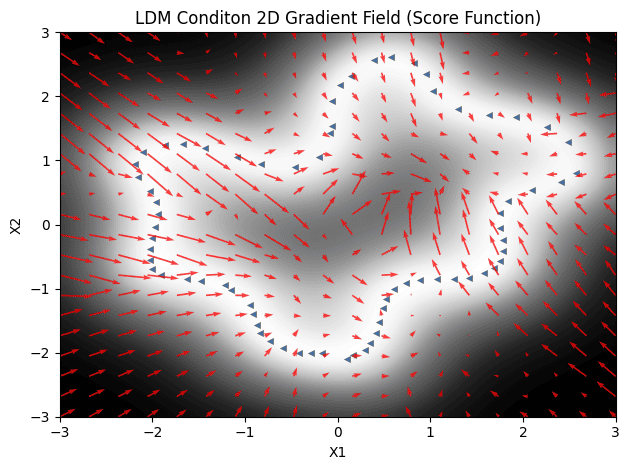}
        \vspace{0.3em}
        \makebox[\linewidth][c]{\footnotesize (b) Condition vector field estimated by LDM}
    \end{minipage}

    \vspace{0.8em}

    \caption
    {\footnotesize
    Comparison of conditional vector fields estimated by ILDM (a) and LDM (b) under the `front view' condition. The blue triangles on the right half of the latent space represent samples from the true conditional distribution corresponding to the front view of the object. In (a), the ILDM vector field (red arrows) consistently points toward these regions, capturing the correct conditional structure. In contrast, (b) shows the LDM vector field, which is disorganized and misaligned with the target distribution.}
    \label{fig:conVec}
\end{figure*}

A comparison of the conditional vector fields produced by ILDM and LDM is presented in Figure~\ref{fig:conVec}. The experiment considers two conditions corresponding to front and back views of the `lucky cat' object. In the latent space, the front-view samples (blue triangles) are concentrated on the right-hand side in Figure~\ref{fig:conVec}(a). As shown, the vector field generated by ILDM (red arrows) consistently points toward this region, indicating precise guidance toward the conditional data distribution. In contrast, the LDM vector field in Figure ~\ref{fig:conVec}(b) exhibits disorganized and inconsistent directions, failing to accurately capture the conditional structure.

When performing backward sampling, classifier-free guidance is implemented by linearly combining the conditional and unconditional score estimates, following the approach of \citet{ho2022classifier} and \citet{salimans2022progressive}
\begin{align}
    \tilde{s}_\theta(x,t,c) = (1 + w) s_\theta(x,t,c) - w s_\theta(x,t),
\end{align}
where the guidance weight 
$w \geq 0$ controls the strength of the conditional signal. When 
$w=0$, the model performs unconditional generation. As demonstrated in prior work \citep{ho2022classifier}, guidance strengths of 
$w\geq4$ typically yield strong guide. Detailed algorithms for classifier free guidance sampling process and conditional generated images are provided in Appendix C.

\section{Experiments}
\label{sec:res}
We evaluate the proposed ILDM on three benchmark datasets: the ‘lucky cat’ object from COIL-100~\citep{nene1996}, a subset of MNIST~\citep{lecun1998mnist}, and cardiac MRI scans~\citep{bernard2018deep,campello2020multi}. These datasets vary in structure and dimensionality, allowing us to assess ILDM’s ability to adapt to different underlying geometries. We compare ILDM against standard diffusion models (DM) and latent diffusion models (LDM), using both qualitative reconstructions and quantitative metrics: Fréchet Inception Distance (FID) and Learned Perceptual Image Patch Similarity (LPIPS)\citep{heusel2017gans, zhang2018unreasonable}. For completeness, the specific SDEs used to implement LDM and DM baselines are detailed in Appendix I.

\subsection{COIL-100 Images}
We first evaluate the proposed ILDM with approximate denoising score matching on the Lucky Cat object from the COIL-100 dataset. The dataset contains 72 color images of the object, each with a resolution of 128×128 pixels and captured at 5° increments to span a complete 360° view. A GP decoder (shown in appendix H-A) 
is pretrained to embed the images to a two-dimensional latent space which defines the Riemannian metric used for diffusion.

\begin{figure}[h]
    \centering
    \subfloat[Original images\label{fig:coil-input-16}]{
        \includegraphics[width=0.45\columnwidth]{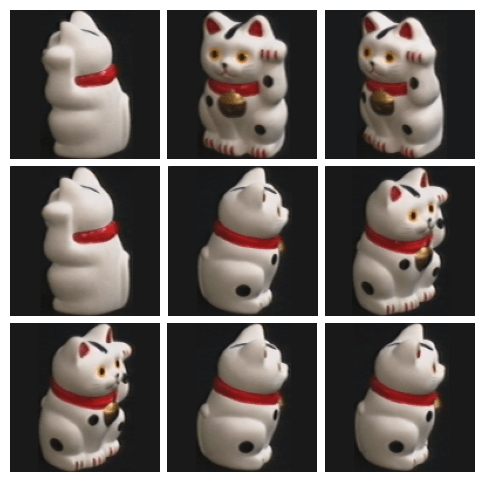}
    }
    \hfill
    \subfloat[ILDM reconstruction\label{fig:coil-rdm-16}]{
        \includegraphics[width=0.45\columnwidth]{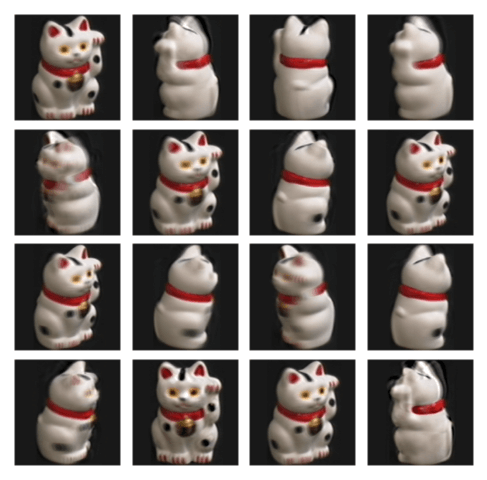}
    }

    \vspace{0.4cm}

    \subfloat[LDM reconstruction\label{fig:coil-ldm-16}]{
        \includegraphics[width=0.45\columnwidth]{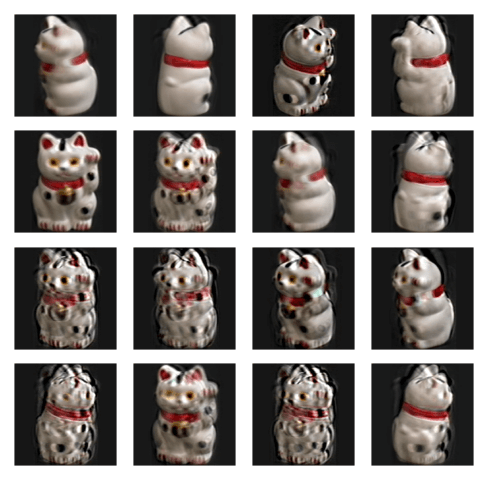}
    }
    \hfill
    \subfloat[DM reconstruction\label{fig:coil-dm-16}]{
        \includegraphics[width=0.45\columnwidth]{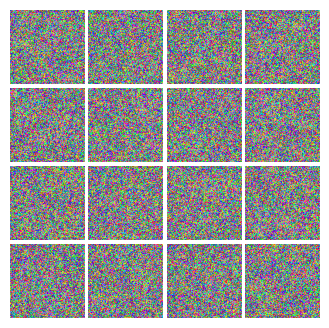}
    }

    \caption{
        \footnotesize{
        Comparison of COIL-100 Lucky Cat image reconstructions using different methods.
        (a) Original input images. (b) Images generated by ILDM with approximate DSM.
        (c) Images generated by the LDM method. (d) Images generated by the DM method.
        }
    }
    \label{fig:coilAll-part}
\end{figure}

Figure~\ref{fig:coilAll-part} provides a qualitative comparison of image reconstructions. Figure~\ref{fig:coilAll-part}(a) shows a subset of training images. Figure~\ref{fig:coilAll-part}(b) presents samples generated by ILDM with  the approximate DSM objective. The reconstructions preserve the global structure, shape, and color consistency of the original object across multiple views, indicating that ILDM successfully captures the data distribution on the unknown manifold. In comparison, Figure~\ref{fig:coilAll-part}(c) presents samples generated by a standard
LDM that performs diffusion in a Euclidean latent space without geometric awareness. These samples exhibit visible distortions and artifacts, particularly in finer details. Figure~\ref{fig:coilAll-part}(d) shows samples generated by score based diffusion model, which largely fail to produce meaningful structure and resemble noise. More generated images can be found in the Appendix D. 

\begin{table}[h]
\caption{Performance comparison of diffusion models on the COIL image dataset. 
Evaluation metrics include FID and LPIPS, where lower values indicate better image quality and perceptual similarity.\label{tab:coil_metrics}}
\centering
\begin{tabular}{|c||c|c|c|}
\hline
\textbf{Method} & \textbf{ILDM} & \textbf{LDM} & \textbf{DM}\\
\hline
FID$\downarrow$ & 154.12 & 171.07 & 352.17\\
\hline
LPIPS$\downarrow$ & 0.5049 & 0.5442 & 1.2905\\
\hline
\end{tabular}
\end{table}

We further assess model performance using quantitative measures. Table~\ref{tab:coil_metrics} summarizes results for three models: ILDM, standard LDM, and a baseline image-space diffusion model (DM). All models generate 200 samples, which are evaluated using the Fréchet inception distance (FID) and learned perceptual image patch similarity (LPIPS). Lower FID and LPIPS scores indicate better visual fidelity and perceptual similarity, respectively. Among the three models, the ILDM achieves the best overall performance, obtaining the lowest FID (154.12) and LPIPS (0.5049) scores. 
The LDM performs moderately well but suffers from degraded perceptual quality.
The baseline DM yields significantly higher FID and LPIPS values, indicating less effective image synthesis under the same evaluation. Additional results for class-conditioned generation are provided in Appendix C.

\begin{figure}[h]
    \centering
    \subfloat[Original images\label{fig:heart-input}]{
        \includegraphics[width=0.45\columnwidth]{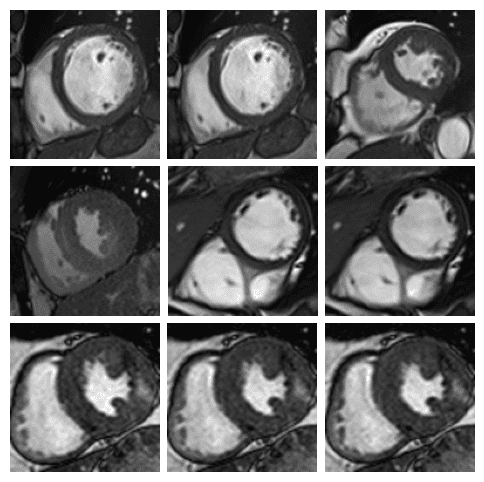}
    }
    \hfill
    \subfloat[ILDM reconstruction\label{fig:heart-rdm-16}]{
        \includegraphics[width=0.45\columnwidth]{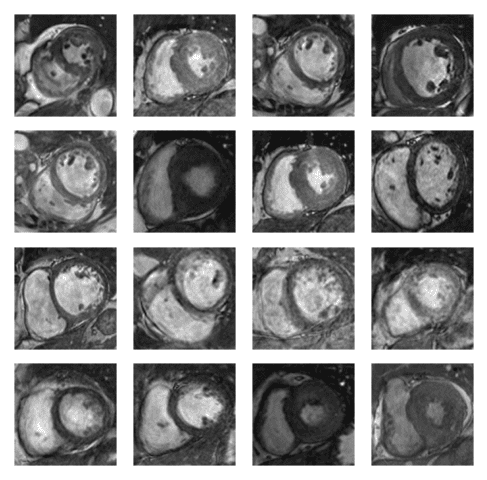}
    }

    \vspace{0.4cm}

    \subfloat[LDM reconstruction\label{fig:heart-ldm-16}]{
        \includegraphics[width=0.45\columnwidth]{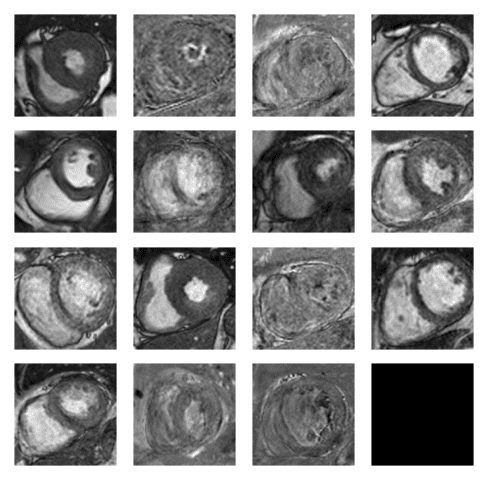}
    }
    \hfill
    \subfloat[DM reconstruction\label{fig:heart-dm-16}]{
        \includegraphics[width=0.45\columnwidth]{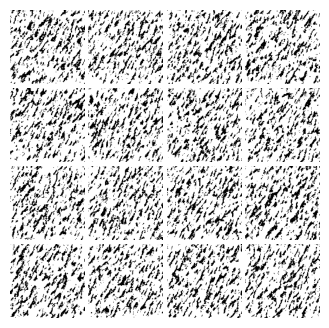}
    }

    \caption{
        \footnotesize{
        Comparison of cardiac MRI image reconstructions using different generative models.
        (a) Original input images. (b) Images generated by the proposed ILDM method.
        (c) Images generated by LDM. (d) Images generated by DM.
        }
    }
    \label{fig:heartAll-part}
\end{figure}

\subsection{ Left Ventricle Cardiac MRI}

We then evaluate the proposed ILDM on a cardiac magnetic resonance image (MRI) dataset focused on the left ventricle. This dataset is constructed from two publicly available sources: the Automatic Cardiac Diagnosis Challenge (ACDC) and the MICCAI 2020 Left Ventricle Challenge \citep{bernard2018deep, campello2020multi}. The dataset comprises 110 healthy subjects and 100 subjects diagnosed with hypertrophic cardiomyopathy (HCM). Each subject contributes two short-axis MRI slices, yielding a total of 330 images. All images were resampled to a resolution of 128 $\times$ 128 pixels. Each image was cropped around the region of interest, specifically focusing on the heart, as illustrated in Figure \ref{fig:heartAll-part}(a). The preprocessed images were then encoded into a five-dimensional latent space for subsequent diffusion.

Figure \ref{fig:heartAll-part}(b) displays images generated by the ILDM method, which largely preserves the overall shape, features, and clear contours. Figure \ref{fig:heartAll-part}(c) shows the result generated by the LDM method, in which the contours are blurred and the shapes do not form clearly. This comparison underscores the enhanced representational and generative abilities of the ILDM method in reconstruction tasks. Additional generated heart images can be found in the Appendix E. 

\begin{table}[H]
\caption{Performance comparison on the left ventricle cardiac MRI dataset.
Evaluation metrics include FID and LPIPS, where lower values indicate better image quality and perceptual similarity.\label{tab:heart-metric}}
\centering
\begin{tabular}{|c||c|c|c|}
\hline
\textbf{Method} & \textbf{ILDM} & \textbf{LDM} & \textbf{DM}\\
\hline
FID$\downarrow$ & 162.94 & 221.88 & 404.53\\
\hline
LPIPS$\downarrow$ & 0.5118 & 0.5278 & 0.7241\\
\hline
\end{tabular}
\end{table}

Table \ref{tab:heart-metric} presents the quantitative results. ILDM achieves the best overall performance, yielding the lowest FID score (162.94) and LPIPS score (0.5118). These results highlight ILDM’s ability to more accurately model the structural and visual characteristics of cardiac MRI data compared to LDM and DM.

\subsection{MNIST}
We further evaluate our method on a subset of the MNIST dataset \citep{lecun1998mnist}, a widely used benchmark composed of grayscale images of handwritten digits (0–9), each with a resolution of 28×28 pixels. For training purposes, we specifically extracted 120 images corresponding to digits 3, 8, and 9. To reflect the intrinsic dimensionality of the underlying manifold, as estimated in prior work \cite{pope2021intrinsic}, we encoded the input images into a 10-dimensional latent space. Representative examples of the selected input images are illustrated in Figure \ref{fig:mnist-input}:

\begin{figure}[h]
    \centering
    \subfloat[Original images\label{fig:mnist-input}]{
        \includegraphics[width=0.45\columnwidth]{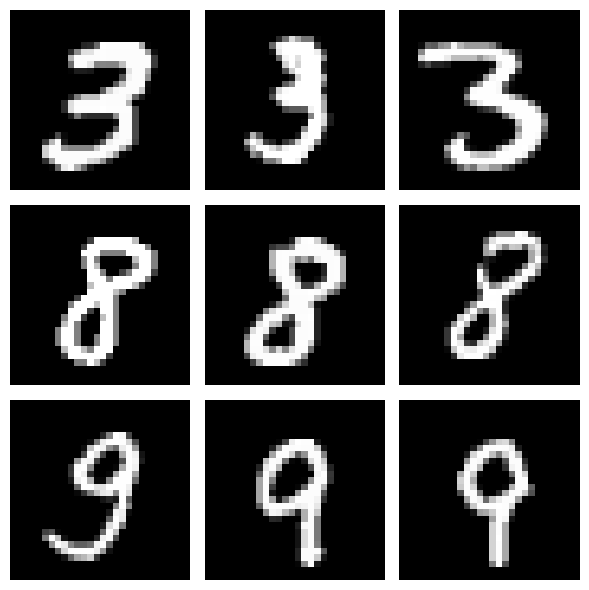}
    }
    \hfill
    \subfloat[ILDM reconstruction\label{fig:mnist-rdm-16}]{
        \includegraphics[width=0.45\columnwidth]{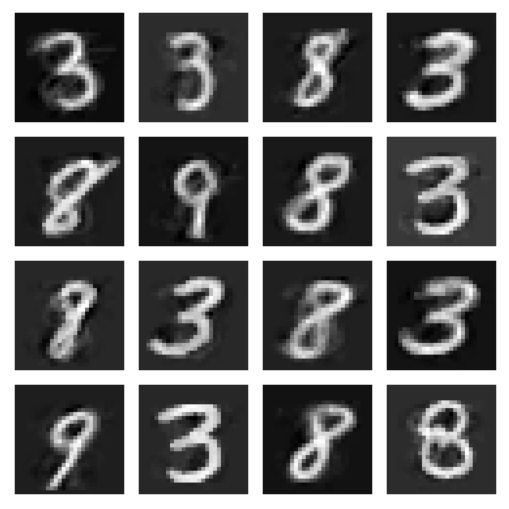}
    }

    \vspace{0.4cm}

    \subfloat[LDM reconstruction\label{fig:mnist-ldm-16}]{
        \includegraphics[width=0.45\columnwidth]{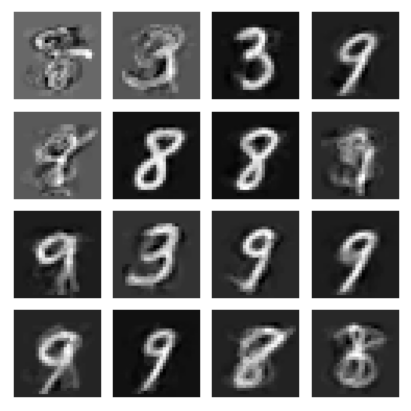}
    }
    \hfill
    \subfloat[DM reconstruction\label{fig:mnist-dm-16}]{
        \includegraphics[width=0.45\columnwidth]{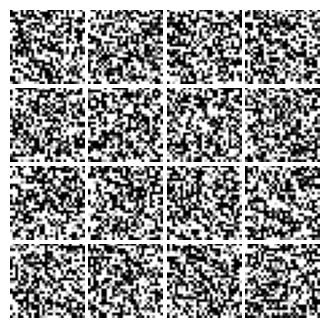}
    }

    \caption{
        \footnotesize{
        Comparison of MNIST image reconstructions using different generative models.
        (a) Original input images. (b) Images generated by the ILDM method.
        (c) Images generated by the LDM. (d) Images generated by the DM.
        }
    }
    \label{fig:mnist-combined}
\end{figure}

Figure \ref{fig:mnist-combined} provides a comparative analysis of MNIST image reconstructions produced by different models. Specifically, the Intrinsic Latent Diffusion Model (ILDM) \ref{fig:mnist-combined}(b) generates reconstructions that are more faithful to the original inputs \ref{fig:mnist-combined}(a), preserving the fine details and overall structure of the digits. Compared to the standard Latent Diffusion Model (LDM) \ref{fig:mnist-combined}(c), the ILDM outputs exhibit clearer contours and more coherent shapes. For a more extensive set of generated MNIST samples, readers are referred to Appendix F. 
We further compare the results quantitatively using FID and LPIPS, as shown in Table \ref{tab:mnist-metric}:

\begin{table}[h]
\caption{Performance comparison on a subset of the MNIST dataset.
Evaluation metrics include FID and LPIPS, where lower values indicate better image quality and perceptual similarity.\label{tab:mnist-metric}}
\centering
\begin{tabular}{|c||c|c|c|}
\hline
\textbf{Method} & \textbf{ILDM} & \textbf{LDM} & \textbf{DM}\\
\hline
FID$\downarrow$ & 136.70 & 155.93 & 353.65\\
\hline
LPIPS$\downarrow$ & 0.6816 & 0.6891 & 0.8892\\
\hline
\end{tabular}
\end{table}

Table~\ref{tab:mnist-metric} provides a quantitative evaluation using FID and LPIPS metrics. The Riemannian LDM achieves the lowest FID score (136.70), outperforming both the standard LDM (155.93) and the baseline image-space DM (353.65). In terms of perceptual similarity, RLDM also performs best, obtaining a slightly lower LPIPS value (0.6816) compared to LDM (0.6891). These results suggest that ILDM better captures the underlying digit structure and perceptual characteristics even in limited-data settings.

\section{Conclusion}
In this work, we propose the Intrinsic Hybrid Latent Diffusion Model (ILDM), a novel generative framework that combines probabilistic dimensionality reduction with score-based diffusion on unknown manifolds. By modeling the latent space as a chart of an data driven Riemannian manifold with geometry induced by a pre-trained mapping function, ILDM captures both the intrinsic geometric structure of the data and the uncertainty of the generative mapping. This integration enables faithful sampling through a hybrid diffusion process that adaptively transitions between Riemannian and Euclidean dynamics based on mapping uncertainty. Empirical evaluations on COIL-100, MNIST, and cardiac MRI datasets show that ILDM consistently outperforms standard diffusion and latent diffusion models in perceptual quality (LPIPS) and fidelity (FID). These findings demonstrate how geometry-aware and uncertainty-adaptive modeling can enhance generative performance, especially when training data is limited.  Future directions include extending ILDM to more expressive decoder architectures, such as attention-based models, and developing more computationally efficient methods for hybrid diffusion sampling and score estimation.

\newpage

\bibliographystyle{IEEEtranN}
\bibliography{new_sample4}

\newpage





\vfill

\end{document}